\documentclass[lettersize,journal]{IEEEtran}
\usepackage{amsmath,amsfonts}
\usepackage{algorithm}
\usepackage{array}
\usepackage[caption=false,font=normalsize,labelfont=sf,textfont=sf]{subfig}
\usepackage{textcomp}
\usepackage{stfloats}
\usepackage{url}
\usepackage{verbatim}
\usepackage{graphicx}
\usepackage{cite}
\usepackage{multicol}
\usepackage{algpseudocode}
\usepackage{multirow}
\usepackage{makecell}
\usepackage{tipa}
\usepackage{balance}
\begin{document}

\title{PhyMamba: Physics-Modulated Mamba for Robust Battery Health Prognostics}

% \author{IEEE Publication Technology,~\IEEEmembership{Staff,~IEEE,}
        % <-this % stops a space

\author{Sara Sameer, \IEEEmembership{Student Member,~IEEE}, Yunyi Zhao, \IEEEmembership{Student Member,~IEEE}, Wei Zhang, \IEEEmembership{Member,~IEEE},\\
Minggang Zeng, Wenqing Li, Man-Fai Ng and Yonggang Wen, \IEEEmembership{Fellow,~IEEE}

\thanks{Manuscript received August 30, 2026; revised XXX, 2026; accepted XXX, 2026. This work was supported in part by A*STAR under its MTC Programmatic (Award M23L9b0052) and MTC Individual Research Grants (IRG) (Award M23M6c0113). (\textit{Corresponding author: Wei Zhang, e-mail: wei.zhang@singaporetech.edu.sg})}
\thanks{Sara Sameer, Yunyi Zhao, Wei Zhang are with the Information and Communications Technology Cluster, Singapore Institute of Technology, Singapore 138683. Yunyi Zhao is also with the Department of Electrical and Computer Engineering, National University of Singapore, Singapore 119077. Minggang Zeng, Wenqing Li, Man-Fai Ng are with the Institute of High Performance Computing (IHPC), Agency for Science, Technology and Research (A*STAR), Singapore 138632. Yonggang Wen is with Nanyang Technological University (NTU), Singapore 639798.}}

% \thanks{This paper was produced by the IEEE Publication Technology Group. They are in Piscataway, NJ.}% <-this % stops a space

% The paper headers
\markboth{IEEE Internet of Things Journal,~Vol.~XX, No.~X, August~2026}%
{Shell \MakeLowercase{\textit{et al.}}: A Sample Article Using IEEEtran.cls for IEEE Journals}

% \IEEEpubid{0000--0000/00\$00.00~\copyright~2021 IEEE}
% Remember, if you use this you must call \IEEEpubidadjcol in the second
% column for its text to clear the IEEEpubid mark.

\maketitle

\begin{abstract}
Battery health prognostics is a core function in battery management systems (BMSs), yet long-horizon health forecasting from BMS signals remains challenging due to operating-condition dependency and sensor noise. In this paper, we propose PhyMamba, a two-stage \underline{phy}sics-modulated \underline{Mamba} framework that integrates electrochemical aging into sequence modelling. PhyMamba does not require explicit identification of internal aging parameters, which often relies on intrusive measurements. In stage-1, a lightweight Mamba encoder first processes BMS signals and produces a latent representation that is transformed via an aging parameterization module, into physics-informed aging features. In stage-2, a customized Mamba forecasting backbone performs multi-cycle prediction, where physics is tightly integrated to regulate the model’s internal temporal updates toward degradation-consistent evolution. Experiments on three public datasets under multiple forecast horizons show that PhyMamba achieves the best aggregated performance, with an overall mean error reduction of 31.8\% compared with a diverse range of baselines. PhyMamba also offers an optimized accuracy-efficiency trade-off, which supports practical deployment for robust battery health prognostics.

\end{abstract}


\section{Introduction}
\label{sec:intro}
Batteries have become a terawatt-hour scale industry. The International Energy Agency (IEA) reports that global battery demand for the energy sector reached the 1 TWh milestone in 2024 \cite{IEA25}. This reflects the rapid scaling of battery usage across transportation and stationary storage and raises the importance of reliable battery monitoring and analytics in real deployments. With such a big scale, even small improvement in batteries' monitoring and management can have large aggregate impacts on safety, efficiency, etc. In operational systems, this is largely mediated by the the battery management system (BMS), which continuously measures battery signals such as voltage and current and produces key state estimates. 
A core BMS function is battery health prognostics, which enables timely protection actions and maintenance planning by modelling and predicting performance degradation \cite{bokstaller2023estimating,bokstaller2024battery}. Degradation is quantified by health indicators, among which a widely used one is the state-of-health (SoH), defined as the ratio between the current usable capacity and the rated capacity. A battery is considered to reach end-of-life (EoL) when its SoH drops below a threshold like 80\%.

Related works on battery health prognostics in the past decade have largely been data-driven with machine learning (ML) algorithms. Many ML models for batteries are purely data-driven, learning end-to-end mappings from BMS signals or their derived features to health indicators. Representative designs include basic regression algorithms \cite{severson2019data}, long short-term memory (LSTM) networks \cite{zhang2018long}, etc. Recently, long-sequence backbones such as Transformer-based architectures \cite{fauzi2024state,sun2025state}, diffusion models \cite{eivazi2024diffbatt,kwan5871665knowdiff} and time-frequency neuromorphic models \cite{ji2024time} have been explored to improve representation capacity and forecasting performance on public benchmarks. In parallel, state-space models (SSMs), especially selective SSMs such as Mamba \cite{gu2024mamba}, have gained attention due to their linear-time sequence modelling and competitive performance in SoH estimation \cite{olalde2024sambamixer}. Robust battery prognostics additionally explain measurement anomalies and outliers, which can substantially distort health-feature extraction and subsequent SoH predictions \cite{wang2026novel}. Meanwhile, recent IoT-oriented battery health models have also emphasized lightweight and edge-efficient inference, highlighting the importance of leveraging prognostic accuracy and computational cost for practical applications \cite{kim2026edge}. Despite these advances, purely data-driven models remain fundamentally constrained by the operating-condition dependency of batteries’ raw measurements, and they can overfit dataset-specific correlations in the measurements rather than capturing degradation-relevant dynamics that transfer across different operating conditions. This motivates the integration of domain knowledge such as battery physics into ML models for battery analytics.

Different forms of domain knowledge can be exploited for battery prognostics. First, several models are developed based on empirical knowledge, e.g., low-frequency signals can exhibit stronger correlation with long-term degradation patterns and degradation typically accelerates as the battery approaches EoL \cite{pamshetti2026knowledge}. Beyond empirical knowledge, domain knowledge from different disciplines has also been incorporated into battery analytics. Equivalent circuit models (ECMs) approximate battery behavior with circuit-level abstractions and estimate circuit states or parameters from BMS signals. ECMs provide interpretable representations with relatively low identification cost, and have been integrated with ML backbones to improve prognostics accuracy \cite{sameer2025pace}. Besides, physics-informed learning injects domain knowledge into ML models through architectural priors and auxiliary constraints to improve prognostic robustness. For example, there is a strong coupling between battery degradation and thermal dynamics and thermal-based entropy is introduced to enhance ML learning \cite{li2026entrolnn}. In addition, electrochemical models are grounded in material-science principles and employ high-fidelity mechanistic formulations such as pseudo-two-dimensional models to capture internal transport and reaction dynamics. However, extracting electrochemical parameters and integrating them in real time are often challenging, as required measurements can be intrusive and computationally expensive. As a result, electrochemical analysis is commonly performed offline under highly calibrated settings \cite{chun2020real}, and the integration of such knowledge into ML remains limited. Overall, existing approaches still face a gap between physics fidelity and deployment scalability. This motivates new models that can utilize electrochemical knowledge to achieve efficient forecasting.

In this paper, our key idea is to leverage electrochemical battery aging physics as the guiding signal for SoH forecasting. A common practice for integrating physics with ML is to augment the model input with physics features, e.g., ECM-derived features in \cite{sameer2025pace}. However, this strategy is less suitable here because electrochemical aging parameters are internal, complex, and challenging to be explicitly and accurately constructed from BMS signals. We therefore adopt a tighter integration. Instead of treating physics aging parameters as additional input, we use physics to regulate a model’s internal process, so that the learned evolution over cycles is driven by degradation-consistent transitions. This becomes particularly natural when we integrate physics with Mamba, whose selective SSM provides an explicit state-evolution mechanism that can be modulated.  

Building on this idea, we in this paper propose PhyMamba, a two-stage \underline{phy}sics-modulated \underline{Mamba} framework for multi-cycle SoH forecasting. In the first stage, a lightweight Mamba encoder processes cycle-indexed BMS logs and learns a latent representation, which is then mapped through an aging-parameterization module into a physics-informed aging feature vector. This design avoids explicit electrochemical parameter identification and instead learns a physically meaningful representation directly from operational data. In the second stage, we perform horizon-based SoH forecasting using a Mamba backbone conditioned on the physics-informed features. Crucially, physics is not only used for feature conditioning, but is also injected into the selective SSM discretization by modulating the step-size term, so that the model's internal update rate adapts to degradation intensity. Together, these designs enable efficient long-horizon forecasting while improving robustness to operating-condition shifts and measurement noise. We conduct extensive experiments to validate the performance of PhyMamba for multi-cycle SoH forecasting across three public datasets and different forecast horizons. Overall, PhyMamba achieves the best performance with the lowest errors with an average of 31.8\% reduction of mean absolute errors (MAEs) and maintains competitive performance across different settings compared against battery models in different ML categories. PhyMamba is also deployment-friendly with a balance between errors and model sizes. 
% (1/8) * [ (1 − 1.262/2.502) + (1 − 1.262/1.990) + (1 − 1.262/1.647) + (1 − 1.262/1.794) + (1 − 1.262/2.037) + (1 − 1.262/2.289) + (1 − 1.262/1.520) + (1 − 1.262/1.489) ] = 31.8%

The remainder of this paper is organized as follows. We first we present related work in Section \ref{sec:review} and the workflow and technical details of PhyMamba in Section \ref{sec:method}. Section \ref{sec:experiment} provides experimental results and discussions. Finally, Section \ref{sec:conclusion} concludes the paper.

\section{Related Work and Preliminaries} \label{sec:review}
This section reviews the SoH forecasting problem, physics-guided battery prognostics, SSM, and Mamba.

% 1. A general literature review
% 2. preliminaries about mamba, how it contributes to our system
% 3. how important is physics informed mamba

\subsection{Physics-Guided Learning for Battery SoH Forecasting}
Accurate battery SoH forecasting is a fundamental task in battery prognostics and health management, with direct implications for safety, reliability, and lifecycle optimization. Existing methods can be broadly categorized into model-based, data-driven, and hybrid physics-informed approaches. Model-based methods describe battery degradation using equivalent-circuit models, empirical aging models, electrochemical models, etc. For example, particle-filtering-based prognostics have been used to model lithium-ion battery capacity depletion and predict remaining useful life (RUL) under uncertainty \cite{saha2009modeling}. Such methods provide physical interpretability and can explicitly reflect degradation mechanisms such as capacity fade, internal resistance growth, and solid electrolyte interphase (SEI) formation. However, their practical deployment is often limited by the need for accurate parameter identification and detailed electrochemical measurements that are not always available from standard BMS logs \cite{berecibar2016critical,xiong2018towards}.

Data-driven methods address these limitations by learning degradation patterns directly from operational measurements, such as voltage, current, and temperature. Classical ML models and deep neural networks have been widely studied for SoH estimation and RUL prediction. Gaussian-process regression has been explored for battery SoH forecasting because it can model nonlinear degradation trends while providing uncertainty estimates~\cite{richardson2017gaussian}. Large-scale early-cycle prediction studies further show that informative features extracted from the first few cycles can support accurate lifetime prediction before obvious capacity degradation occurs~\cite{severson2019data}. Review studies have also summarized the rapid development of data-driven battery health estimation and lifetime prediction, covering feature engineering, deep learning, and real-time deployment challenges~\cite{li2019data}. Although these approaches are effective in extracting nonlinear temporal patterns from BMS data, their predictions may be sensitive to operating-condition shifts and limited labeled data. Moreover, purely data-driven models often provide limited insight into the underlying degradation mechanisms, making it difficult to assess whether the learned temporal dependencies are physically meaningful.

Physics-informed learning has therefore become an important direction for battery health prognostics. Instead of relying solely on black-box statistical correlations, physics-informed methods incorporate degradation knowledge through constrained loss functions, physically meaningful latent variables, etc. Hybrid physics-informed neural networks have been proposed to combine reduced-order battery models with neural networks for battery modeling and prognosis~\cite{nascimento2021hybrid}. Physics-informed ML has also been used for SoH prognostics from partial charging segments by embedding SEI-growth-related degradation knowledge into the learning process~\cite{kohtz2022physics}. Other studies integrate empirical or semi-physical degradation models with neural networks to fuse prior degradation dynamics and monitoring data for SoH and RUL prediction~\cite{wen2023physics}. More recently, physics-informed neural networks have shown strong potential for stable lithium-ion battery degradation modeling and SoH estimation under diverse battery types and operating conditions~\cite{wang2024physics}. These studies suggest that physical knowledge can improve robustness and interpretability in battery health modeling.

However, many existing physics-informed approaches still use physics mainly as an external regularizer or auxiliary loss. In contrast, the goal of our work is to embed aging physics into the internal temporal update mechanism of the sequence model. This enables the learned memory dynamics to be directly regulated by the estimated degradation state, rather than treating physical information as a passive input appended to the data-driven predictor.

\subsection{SSMs for Long-Sequence Modeling}
SSMs have recently emerged as a promising alternative to attention-based architectures for long-sequence modeling. Classical SSMs represent sequence dynamics through a latent state that evolves over time. In this formulation, the latent state acts as a compact memory of past inputs, while the state transition and skip connection determine how information is retained and propagated. This structure makes SSMs naturally suitable for continuous signals and time series with long-range temporal dependencies.
Modern deep SSMs extend this classical formulation by learning state-space parameters within neural sequence models. A representative milestone is S4, which introduced efficient structured parameterizations that made SSMs practical for long sequences while preserving both recurrent and convolutional interpretations~\cite{gu2021s4}. S4 demonstrated that SSMs could compete with Transformer-based architectures on tasks requiring long-range dependency modeling. Subsequent works further simplified and improved SSM architectures. DSS showed that diagonal SSMs can be competitive with more structured SSMs~\cite{gupta2022dss} and S4D studied the parameterization and initialization of DSS~\cite{gu2022s4d}. S5 introduced a simplified multi-input multi-output SSM layer with efficient parallel-scan computation~\cite{smith2022s5}.
%; and H3 explored how SSMs can better support language modeling by improving recall and comparison mechanisms~\cite{fu2023h3}. 
These developments show a clear progression from fixed long-range sequence models toward more expressive, scalable, and hardware-efficient SSM architectures.

%For battery prognostics, SSMs are particularly attractive because battery degradation is inherently sequential and often evolves over long cycle histories. The latent state in an SSM can summarize historical degradation information, while the transition and update mechanisms determine how strongly past and current observations influence future predictions. This provides a natural modeling framework for SoH forecasting, where the relevance of historical cycles may vary across different degradation regimes.

%\subsection{Mamba Preliminaries}
Mamba represents a major advancement in the SSM family by introducing selective state space models~\cite{gu2023mamba}. Unlike earlier SSMs whose dynamics are largely fixed after training, Mamba makes several key state-space quantities depend on the current input representation. This input-dependent selection allows the model to decide, at each step, whether to retain previous information, forget stale information, or write new information into the latent state. As a result, Mamba improves the expressiveness of SSMs while retaining their efficient recurrent structure.
Compared with Transformers, which use self-attention to compare tokens across the full context, Mamba compresses historical information into a recurrent hidden state and updates it sequentially. This enables linear scaling with sequence length and avoids the need for a growing key-value cache during auto-regressive inference. These properties make Mamba well suited for long-sequence tasks such as battery health prognostics where efficiency and memory scalability are important. Recent developments such as Mamba-2 further reveal theoretical connections between SSMs and attention through structured state-space duality~\cite{dao2024mamba2}, while hybrid architectures such as Jamba combine Transformer and Mamba layers to leverage the strengths of both attention and SSMs~\cite{lieber2024jamba}. This suggests that SSMs are becoming practical components of next-generation sequence modeling architectures.

%In our system, Mamba contributes in two complementary ways. First, it serves as an efficient sequence encoder in the aging feature construction stage, where historical BMS measurements are mapped into compact latent electrochemical descriptors. Second, it provides the forecasting backbone in the health prediction stage. This second role is especially important because Mamba exposes interpretable state-space components: the transition term controls memory retention, the input term controls how strongly new information updates the latent state, the readout term controls how the latent memory is converted into an output representation, and the skip connection provides a direct residual pathway. These components create a natural interface for incorporating degradation physics into the temporal modeling process.

% \subsubsection{Importance of Physics-Informed Mamba}
Although Mamba provides an effective data-dependent selection mechanism, standard Mamba remains agnostic to the physical meaning of the sequence. Its step size, input write term, and readout term are learned from data alone. For battery health forecasting, this is potentially limiting because degradation dynamics are not arbitrary temporal patterns. They are governed by electrochemical aging mechanisms whose influence changes across early-life, mid-life, and end-of-life regimes. A purely data-driven selective SSM may capture these effects implicitly, but it may also learn spurious temporal correlations under noisy measurements or changing operating conditions.
To address this limitation, PhyMamba introduces a physics-informed Mamba architecture for battery health forecasting. It first extracts a compact electrochemical aging representation from BMS observations and then uses the aged physics features to condition the Mamba forecasting stage. The key idea is to embed degradation knowledge into the selective SSM dynamics, rather than treating physics as auxiliary input only. As a result, the model retains the efficiency and adaptive selection capability of Mamba while making the temporal update mechanism more consistent with battery aging behavior.

The importance of physics-informed Mamba lies in its ability to connect data-driven temporal modeling with physically meaningful degradation states. In mild aging regimes, battery degradation is usually slow and smooth, so longer memory over historical cycles can be beneficial. In severe aging regimes, degradation may accelerate, making recent observations more informative for future SoH prediction. By allowing the aging representation to regulate the internal update behavior of the Mamba backbone, PhyMamba enables the model to adapt its memory and update strength according to the estimated degradation condition.
This design also improves interpretability. The transition component of the SSM can be understood as controlling the degradation-dependent memory horizon, the input component controls how strongly new degradation information is written into the latent state, the readout component adapts the output representation to different degradation compositions, and the skip pathway allows aged electrochemical features to directly contribute to the prediction. Therefore, physics-informed Mamba is not only a feature-enhanced sequence model, but a mechanism-level integration of electrochemical aging knowledge and selective SSM. This makes it particularly suitable for battery health forecasting, where accurate prediction requires both long-context temporal modeling and consistency with physical degradation.

\section{PhyMamba: Methodology}
\label{sec:method}
This section presents the proposed PhyMamba framework, including the problem formulation, physics-informed aging feature construction, and physics-modulated forecasting.

\subsection{Battery Health Prognosis and System Overview}

%\subsubsection{Background and Key Challenges}
Battery health is commonly learned from time-series sensor measurements collected by BMS. In practice, battery health models take as input either raw measurements from multiple charge/discharge cycles or derived per-cycle features, and are often supervised using historical health targets such as capacity or SoH. Since degradation accumulates over time, prognostics models are typically sequential, where the models process historical data to estimate current health and predict future degradation trends. 
Existing models are predominantly data-driven. However, battery degradation data presents several characteristics that make robust learning difficult. First, the measured trajectories are highly dependent on operating conditions, such as charging/discharging protocols, ambient temperature, and load, which can lead to distribution shifts between model training and deployment. Second, the measurements can be non-stationary and noisy, with transient fluctuations or occasional spikes caused by sensing noise and short-term operational disturbances, e.g., regenerative braking during vehicle driving \cite{li2025driving}. These effects may bias sequence models toward short-term variations rather than long-term degradation evolution. Third, many ML models are computationally intensive; however, degradation accumulates over long horizons, which requires the models to capture long-range dependencies for effective prognostics while remaining computationally efficient for long data sequences. These challenges require new models that are accurate, reliable, and lightweight.

%\subsubsection{Problem Formulation}
Given a battery's BMS signals as measurement time-series, we construct data samples for ML model training at a given current cycle $c$ by extracting a fixed-length history window from the time-series data. Let the input of a model be $X_c \in \mathbb{R}^{T \times d}$, which represents the data of a battery's past $T$ cycles ending at cycle $c$. Here, $T$ denotes the length of the input window in terms of the number of cycles within the window. $d$ is the input dimension, e.g., $d=2$ if the input includes voltage and current time-series only. We write $X_c=\{x_t\}_{t=c-T+1}^{c}$, where $x_t$ is the $d$-dimensional feature vector at cycle $t$ within the input window. When available, auxiliary BMS metadata, such as rated battery capacity and remaining maximum capacity, can be included in $x_t$, which increases the effective feature dimension $d$.
Then, we define a forecast horizon of length $H$ cycles and aim to predict the battery health over the future cycles $\{c+1,\ldots,c+H\}$. We consider a prognostic model $f_{\boldsymbol{\theta}}(\cdot)$ with parameters $\boldsymbol{\theta}$ that maps the input to a sequence of health predictions, i.e., $\{\hat{y}_{c+t}\}_{t=1}^{H} = f_{\boldsymbol{\theta}}(X_c)$, where $\hat{y}_{c+t}$ denotes the predicted health value at cycle $c+t$ and we use SoH as the health indicator. $\boldsymbol{\theta}$ is learned by solving the following optimization problem,
\begin{equation}
\boldsymbol{\theta}^{*}=\arg \min_{\boldsymbol{\theta}}\sum_{i=1}^{n}\sum_{t=1}^{H}\ell \left(\hat{y}_{c+t}^{\,i},\, y_{c+t}^{\,i}\right),
\end{equation}
where $n$ is the number of training data samples and $\ell(\cdot,\cdot)$ is a loss or error function, e.g., mean squared error (MSE). Two terms in the error function correspond to the predicted and the corresponding ground-truth SoH, respectively.

\begin{figure*}[t]
\centering
\includegraphics[width=0.98\textwidth]{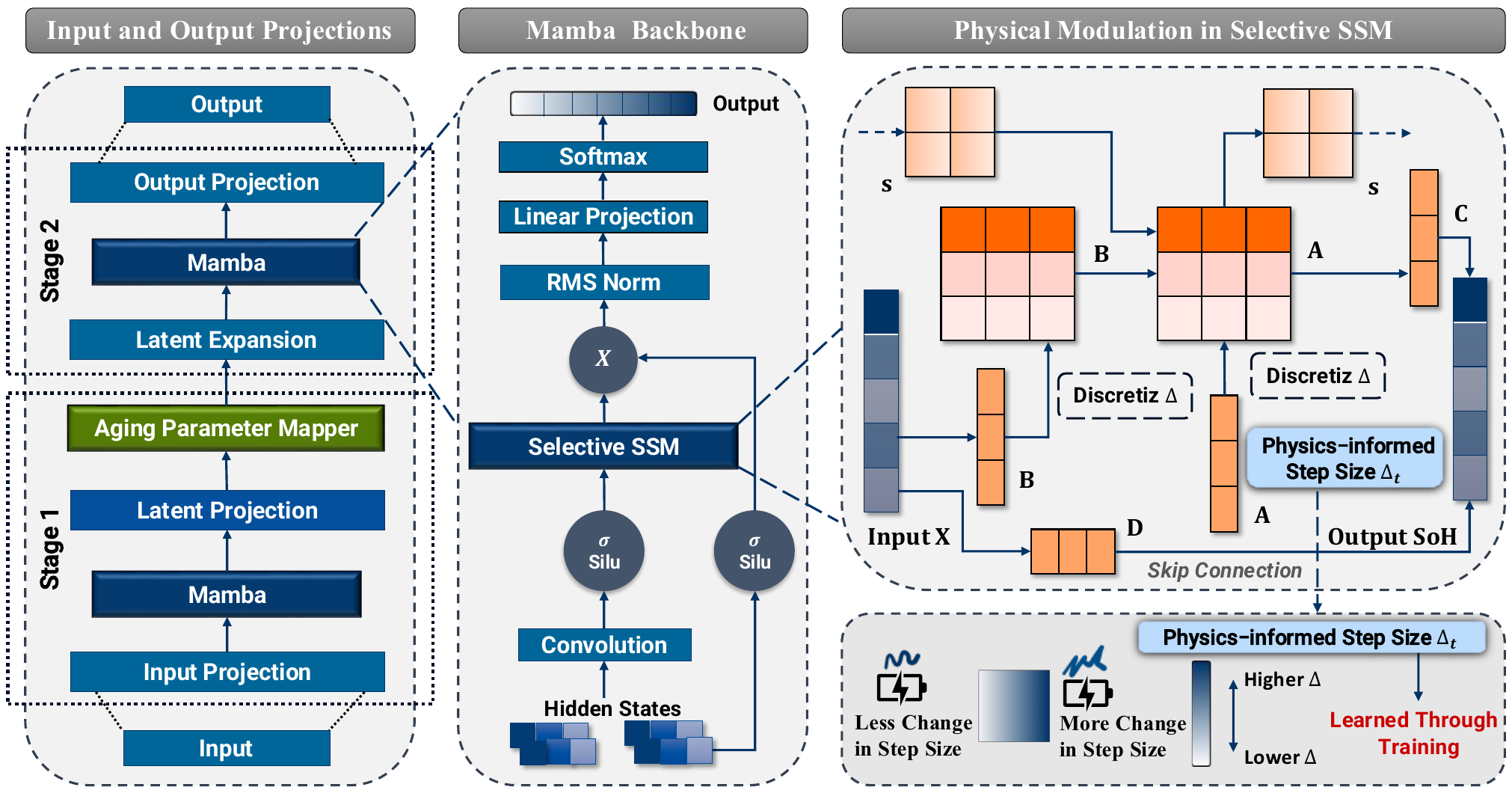}
\caption{An illustration of the two-stage system architecture of PhyMamba. Stage-1 infers physics-informed battery aging parameters from BMS logs. Stage-2 performs multi-cycle SoH forecasting with a physics-modulated Mamba backbone.}
\label{fig:sys}
\end{figure*}

%\subsubsection{PhyMamba Overview}
In this paper, we propose PhyMamba, a two-stage framework that follows the input-output definition in the above formulation. An overview of the system architecture of PhyMamba is shown in Fig. \ref{fig:sys}. Overall, PhyMamba maps the input data to the health predictions in the forecast horizon. To achieve this, PhyMamba first extracts a physics-informed feature representation that is compact yet informative from the input. This representation is designed to capture degradation-relevant physics in a structured manner and is then provided to the next stage, which performs multi-cycle forecasting over the horizon. In the following subsections, we detail our unique designs. 
% To ensure both prognostic scalability and robustness, the forecasting stage is implemented with a Mamba-based selective state-space backbone. The design enables efficient long-sequence modelling, where physics regulates the model's state evolution. 

\subsection{Physics-based Aging Feature Construction}

%\subsubsection{Physics-Informed Aging Parameterization}
Battery degradation is governed by coupled electrochemical, interfacial, and transport processes that evolve gradually during cycling. Although BMS measurements such as voltage and current are readily available, they only provide indirect observations of these internal aging mechanisms. Motivated by established electrochemical aging models and prior scientific studies~\cite{chun2020real}, we summarize the degradation state using a small set of physically interpretable parameters, each associated with a known aging-related property. Following this idea, PhyMamba adopts a compact six-dimensional physics-informed parameterization, which is later evolved into aging features for robust battery health prognostics. For clarity, we group the six parameters into three categories.

%\paragraph{Electrode Properties}
The first category describes electrode properties using four complementary parameters: the cathode and anode active surface areas $a_s^p$ and $a_s^n$, and the cathode and anode effective conductivities $\sigma_s^p$ and $\sigma_s^n$, where superscripts $p$ and $n$ denote the positive and negative electrodes, respectively. In electrochemical modeling, the surface area represents the available electrode interface for lithium intercalation reactions, while the conductivity characterizes electron transport efficiency through the electrode structure. During aging, mechanisms such as particle cracking, loss of active material, surface film growth, and increasing contact resistance can reduce the accessible reaction area and degrade effective conductivity. These changes increase polarization and internal losses, thereby altering the voltage and current trajectories observed by the BMS. Therefore, these four parameters provide a compact physical summary of electrode-property degradation.

%\paragraph{Interfacial Degradation}
The second category captures interfacial degradation through the solid electrolyte interphase thickness, denoted by $L_{\mathrm{SEI}}$. The SEI layer forms on the anode surface and generally grows during cycling due to parasitic side reactions. A thicker SEI layer increases interfacial impedance and consumes cyclable lithium, which contributes to both capacity loss and increased polarization. Since these effects are reflected in terminal battery behavior, $L_{\mathrm{SEI}}$ serves as an informative physics-aligned indicator linking internal aging progression to observable health degradation. 

%\paragraph{Health Reference} 
The third category provides a direct health reference using the normalized capacity, denoted by $C_{\mathrm{norm}}$. Unlike surface area, conductivity, or SEI thickness, $C_{\mathrm{norm}}$ is not a mechanism-specific degradation parameter. Instead, it represents the remaining usable capacity relative to the reference capacity condition. Because capacity fade is a primary outcome of battery aging and is closely related to SoH, $C_{\mathrm{norm}}$ anchors the physics-informed parameterization to the health prediction target and complements the internal degradation descriptors above. 

In this part, we bridge the gap between observable BMS signals and the physics-informed aging parameters introduced above by developing an aging feature construction module, which serves as the first stage of PhyMamba. In practice, electrochemical aging parameters are not directly available from standard BMS logs; quantities such as SEI thickness or effective electrode surface area usually require intrusive or laboratory-level characterization. Therefore, we combine operational BMS data with domain knowledge encoded through the physics-informed parameterization. Given the input sequence $X_c$, the module first learns a compact six-dimensional beginning of life (BoL) latent parameter vector that summarizes degradation-relevant internal properties. This vector is then transformed by a reduced-order aging model into aged electrochemical features, which are used by the second-stage PhyMamba forecaster for multi-cycle health prediction.

We implement the first component as a lightweight Mamba-based sequence encoder over the input $X_c$. Each $d$-dimensional input vector is projected to a hidden dimension $h_1$ using a \texttt{Linear} layer. The resulting hidden sequence is processed by a stack of $L_1$ \texttt{Mamba} layers to capture temporal dependencies in voltage, current, and time-related measurements. A \texttt{Linear} output head then maps the encoded representation to a six-dimensional vector. We denote the inferred BoL parameter by,
\begin{equation}
\boldsymbol{\phi}_{\mathrm{BoL}}
=
f_{\mathrm{enc}}(X_c)
\in \mathbb{R}^{6}.
\end{equation}
Here, $\boldsymbol{\phi}_{\mathrm{BoL}}$ is a data-inferred latent vector rather than a directly measured electrochemical state. Its dimensionality is aligned with the six physics parameters described above, so that each channel can be interpreted as the BoL baseline of one aging-related parameter. Although we use a Mamba encoder for consistency with the overall architecture, this component can in principle be replaced by other sequence encoders, such as Transformer-based models.

Next, we introduce an aging parameter mapping module to convert $\boldsymbol{\phi}_{\mathrm{BoL}}$ into a physics-informed aged feature vector. Specifically, we use a reduced-order electrochemical aging equation~\cite{chun2020real}. The key idea is to express each physics channel as its BoL baseline minus an accumulated degradation term driven by SoH change and cycle progression. Let $N_{\mathrm{ap}}^{i}\geq 1$ denote the number of aging phenomena contributing to parameter channel $i$, let $\phi_{\mathrm{BoL}}^{i}$ be its inferred BoL baseline, and let $\Delta_{\mathrm{SoH}}$ denote the normalized SoH change relative to the baseline. We model the aged value of channel $i$ as,
\begin{equation}
\phi_{\mathrm{aging}}^{i} = \phi_{\mathrm{BoL}}^{i} - \sum_{j=1}^{N_{\mathrm{ap}}^{i}}
k_i
\int_{0}^{\Delta_{\mathrm{SoH}}}
\left(F_i-(F_i-1)x\right)^{D_{ij}}\mathrm{d}x ,
\label{eq:aging_integral}
\end{equation}
where $k_i$ and $F_i$ are signed empirical coefficients listed in Table~\ref{tab:para}, and $D_{ij}$ is a learnable occurrence degree controlling the contribution strength of aging phenomenon $j$ to parameter $i$. Importantly, $D_{ij}$ acts as a multiplicative contribution factor rather than an exponent. This allows different aging phenomena to contribute with different strengths while preserving the same physics-motivated degradation form across channels.
% The integral in Eq.~\eqref{eq:aging_integral} has the following closed form:
% \begin{equation}
% \int_0^x \bigl[F_i - (F_i - 1)\,u\bigr]^{D_{ij}}\,du
% = D_{ij}\left(F_i x - \frac{(F_i - 1)\,x^2}{2}\right)
% \label{eq:integral_closed}
% \end{equation}
% When accumulated over $N_{\mathrm{cycles}}$ cycles, the aged parameter can be written as
% \begin{equation}
% \theta_i^{\mathrm{aging}} = \theta_i^{\mathrm{BoL}}
%     - k_i\, D_{ij}\left(F_i\,\mathrm{SoH} - \frac{(F_i-1)\,\mathrm{SoH}^2}{2}\right)
%     \times N_{\mathrm{cycles}}
% \label{eq:total_degradation}
% \end{equation}
Collecting the six updated channels yields the aged electrochemical feature vector,
\begin{equation}
\boldsymbol{\phi}_{\mathrm{aging}}
=
[a_s^p, a_s^n, \sigma_s^p, \sigma_s^n, L_{\mathrm{SEI}}, C_{\mathrm{norm}}]^\top
\in \mathbb{R}^{6}.
\end{equation}
This vector is the physics-informed representation passed to the forecasting stage. In this way, PhyMamba does not require direct measurement of internal electrochemical states, but instead learns BoL-aligned latent parameters from BMS data and evolves them through a differentiable aging model before using them for SoH prediction. Details of the mentioned parameters can be found in Table~\ref{tab:para}.

\begin{table*}[h!]
\centering
\footnotesize
\caption{Summary of physics parameters and aging phenomena coefficients that are specified based on empirical studies and domain knowledge.}
\label{tab:para}
\renewcommand{\arraystretch}{1.2}
\begin{tabular}{c|c|c|c|c|c}
\hline\hline
Category & Parameter & Physical Description & $N_{\text{ap}}$ & $k$ & $F$ \\
\hline\hline
\multirow{4}{*}{Electrode Properties}
& $a_{\mathrm{cp}}$        & cathode surface area        & 2 & +0.2 & 39.80  \\ \cline{2-6}
& $a_{\mathrm{an}}$        & anode surface area          & 2 & +0.2 & $-35.08$ \\ \cline{2-6}
& $\sigma_{\mathrm{cp}}$   & cathode conductivity        & 1 & $-0.2$ & 31.76 \\ \cline{2-6}
& $\sigma_{\mathrm{an}}$   & anode conductivity          & 1 & $-0.2$ & $-62.72$ \\
\hline
Interfacial Degradation
& $L_{\mathrm{SEI}}$   & SEI layer thickness         & 3 & +0.2 & $-50.0$ \\
\hline
Health Reference
& $C_{\mathrm{norm}}$ & normalized capacity      & 1 & +0.2 & 1.0 \\
\hline\hline
\end{tabular}
\end{table*}

\subsection{Forecasting with Physics-Modulated Mamba}
In the second stage of PhyMamba, we perform multi-cycle SoH forecasting using the physics-informed aging features constructed in the first stage. Let
$\boldsymbol{\phi}_c \equiv \boldsymbol{\phi}_{\mathrm{aging},c}\in\mathbb{R}^{6}$
denote the aged electrochemical feature vector at the current cycle $c$. This vector summarizes the estimated degradation state of the battery, including electrode-property degradation, SEI growth, and normalized capacity fade. The forecasting module maps $\boldsymbol{\phi}_c$ to future health predictions
$\{\hat{y}_{c+t}\}_{t=1}^{H}$ over the forecast horizon $H$. Instead of using the physics features only as ordinary input variables, PhyMamba injects them into the internal state-space dynamics of Mamba. Therefore, the degradation state directly regulates how the model retains historical information, writes new information into memory, reads out latent states, and passes physics features to the prediction output.
%\subsubsection{Mamba-based Forecasting Backbone}
The forecasting stage adopts a Mamba-based selective state-space backbone~\cite{gu2023mamba}. The input aging feature vector $\boldsymbol{\phi}_c$ is first projected to a hidden dimension $h_2$ using a \texttt{Linear} layer,
\begin{equation}
    u_t = \mathrm{Proj}_{h_2}(\boldsymbol{\phi}_c),
\end{equation}
where $u_t$ denotes the physics-conditioned hidden representation used by the selective SSM at model step $t$. The hidden representation is then processed by a stack of $L_2$ \texttt{Mamba} layers. Finally, a \texttt{Linear} output head maps the final representation to an $H$-dimensional prediction vector, followed by a sigmoid activation to produce normalized SoH predictions,
\begin{equation}
    \{\hat{y}_{c+t}\}_{t=1}^{H}
    =
    \mathrm{Sigmoid}
    \left(
    \mathrm{Linear}_{H}
    \left(
    \mathrm{Mamba}_{L_2}
    \left(
    \mathrm{Linear}_{h_2}(\boldsymbol{\phi}_c)
    \right)
    \right)
    \right).
\end{equation}
The use of Mamba is motivated by its selective state-space formulation. Unlike attention-based models, whose pairwise token interactions scale quadratically with sequence length, Mamba uses a recurrent state-space update that scales linearly with sequence length while preserving a content-dependent selection mechanism. This is suitable for battery prognostics, where degradation evolves over long cycle histories but the relevance of historical information changes with the current health condition. More importantly, Mamba exposes interpretable state-space components, including transition, input, readout, and residual terms, which provide a natural interface for incorporating electrochemical aging information into the forecasting dynamics.

%\subsubsection{Background: Selective State-Space Models in Mamba}
Mamba is built upon structured SSMs. A continuous-time SSM maps an input signal $x(t)$ to an output signal $y(t)$ through a latent state $s(t)$,
\begin{equation}\label{eq:continuous_ssm}
\begin{split}
    \dot{s}(t) &= A s(t) + B x(t), \\
    % \qquad
    y(t) &= C s(t).
\end{split}
\end{equation}
Here, $A$ is the state transition matrix, $B$ is the input matrix, and $C$ is the output or readout matrix. The latent state $s(t)$ serves as a compact memory of previous inputs. In many practical SSM implementations, including Mamba-style blocks used in PhyMamba, an additional skip or residual term is included as,
\begin{equation}
    y(t) = C s(t) + D x(t),
    \label{eq:ssm_skip}
\end{equation}
where $D$ allows the input to directly contribute to the output without passing through the recurrent state.
For discrete sequence modeling, the continuous dynamics in Eq.~\eqref{eq:continuous_ssm} are converted into a recurrence using a step size $\Delta$. Under zero-order-hold discretization, the continuous parameters $(\Delta,A,B)$ are transformed into discrete parameters $(\bar{A},\bar{B})$,
\begin{equation}
    \bar{A} = \exp(\Delta A),
    \qquad
    \bar{B} = (\Delta A)^{-1}\left(\exp(\Delta A)-I\right)\Delta B.
    \label{eq:zoh_discretization}
\end{equation}
The resulting discrete-time recurrence is
\begin{equation}
    s_t = \bar{A}s_{t-1}+\bar{B}x_t,
    \qquad
    y_t = C s_t.
    \label{eq:discrete_ssm}
\end{equation}
When the parameters are fixed across time, the SSM is linear time-invariant. Such models can be computed either as recurrent updates or as global convolutions. However, fixed dynamics limit the model's ability to selectively keep or discard information depending on the current input.
Mamba addresses this limitation through a selective SSM. Instead of keeping all SSM parameters fixed across the sequence, Mamba makes the step size $\Delta_t$, input matrix $B_t$, and readout matrix $C_t$ depend on the current representation,
\begin{equation}
    B_t = s_B(x_t),
    \qquad
    C_t = s_C(x_t),
    \qquad
    \Delta_t = \tau_{\Delta}\left(\theta_{\Delta}+s_{\Delta}(x_t)\right),
    \label{eq:mamba_selectivity}
\end{equation}
where $s_B(\cdot)$, $s_C(\cdot)$, and $s_{\Delta}(\cdot)$ are lightweight projections, $\theta_{\Delta}$ is a learnable bias, and $\tau_{\Delta}$ is typically implemented as \texttt{Softplus} to ensure that $\Delta_t>0$. This makes the recurrence time-varying,
\begin{equation}
    \bar{A}_t = \exp(\Delta_t A),
    \qquad
    \bar{B}_t \approx \Delta_t B_t,
    \label{eq:mamba_discrete}
\end{equation}
\begin{equation}
    s_t = \bar{A}_t s_{t-1} + \bar{B}_t x_t,
    \qquad
    y_t = C_t s_t + D x_t.
    \label{eq:mamba_recurrence}
\end{equation}
The approximation $\bar{B}_t \approx \Delta_t B_t$ is commonly used in the simplified selective-scan implementation and is also the form adopted in PhyMamba. In this formulation, the role of $\Delta_t$ is especially important: it acts similarly to a learned gate that controls how quickly the latent state evolves. A small $\Delta_t$ encourages memory retention, while a large $\Delta_t$ makes the state update more rapidly and discounts older information.

%\subsubsection{Physics-Modulated Selective SSM Update}
Building on the selective SSM formulation, PhyMamba introduces physics modulation by injecting a physics-informed bias into the step-size pathway. In standard Mamba, $\Delta_t$ is determined by the current hidden representation through a learned projection. In PhyMamba, the current hidden representation $u_t$ originates from the aged electrochemical feature vector $\boldsymbol{\phi}_c$. Therefore, its magnitude reflects the current degradation condition of the battery. PhyMamba computes the step size as,
\begin{equation}
    \Delta_t
    =
    \mathrm{Softplus}
    \left(
    W^{\Delta}\tilde{\delta}_t
    +
    b^{\Delta}
    +
    \alpha |u_t|
    \right),
    \label{eq:physics_delta}
\end{equation}
where $\tilde{\delta}_t$ is the unconstrained step-size score produced by the standard Mamba pathway, $W^{\Delta}$ and $b^{\Delta}$ are learnable calibration parameters, $\alpha$ is a learnable per-channel physics-modulation coefficient, and $|u_t|$ denotes the element-wise absolute value. The first two terms,
$W^{\Delta}\tilde{\delta}_t+b^{\Delta}$, correspond to the original Mamba step-size pathway, while the additional term $\alpha |u_t|$ is the physics-informed bias introduced by PhyMamba.
This design has a direct theoretical effect on the discretized transition matrix. In Mamba-style SSMs, the continuous transition parameter is usually constrained to be negative, for example by,
\begin{equation}
    A = -\exp(A_{\log}),
    \label{eq:negative_A}
\end{equation}
which ensures stable decay dynamics. Substituting Eq.~\eqref{eq:physics_delta} into the discretized transition gives,
\begin{equation}
    \bar{A}_t
    =
    \exp(\Delta_t A),
    \qquad
    A<0.
    \label{eq:physics_A}
\end{equation}
Since $A$ is negative, $\bar{A}_t$ lies between $0$ and $1$. Specifically, $\Delta_t$ approaches to $0$ when $\bar{A}_t$ becomes $1$ and conversely $\Delta_t \rightarrow \infty$ when $\bar{A}_t$ is close to $0$. Such behavior corresponds to strong retention of the previous state and fast forgetting of previous memory.
% \begin{equation}
%     \Delta_t \rightarrow 0
%     \quad \Rightarrow \quad
%     \bar{A}_t \rightarrow 1,
% \end{equation}
% which corresponds to strong retention of the previous state, while
% \begin{equation}
%     \Delta_t \rightarrow \infty
%     \quad \Rightarrow \quad
%     \bar{A}_t \rightarrow 0,
% \end{equation}
% which corresponds to fast forgetting of previous memory.

Physically, when the battery is in an early or mild aging regime, the magnitude of the aged feature representation is relatively small, so the physics bias remains weak and the model preserves longer-term degradation memory. When degradation becomes severe, such as under significant SEI growth or loss of active electrode surface area, the term $\alpha |u_t|$ increases $\Delta_t$, causing the model to discount older states and focus more on recent cycles. This matches the accelerating nature of late-stage battery degradation, where recent measurements often become more informative for future SoH.
The same physics-modulated step size also affects the input matrix. In the selective SSM update,
\begin{equation}
    \bar{B}_t = \Delta_t B_t,
    \label{eq:physics_B}
\end{equation}
where $B_t=s_B(u_t)$ is already input-dependent. Therefore, the physics term influences $B$ in two ways. First, the input projection $B_t$ is computed from the physics-conditioned representation $u_t$. Second, the effective write strength $\bar{B}_t$ is scaled by the physics-modulated step size $\Delta_t$. As a result, stronger degradation signals increase the contribution of the current representation to the latent state. When $\alpha|u_t|$ increases, $\Delta_t$ and $\bar{B}_t$ will become larger simultaneously. This gives PhyMamba an adaptive write mechanism. When under high electrochemical stress, each new cycle can update the latent memory more strongly, while under mild degradation the model evolves more smoothly.
% \begin{equation}
%     \alpha|u_t| \uparrow
%     \quad \Rightarrow \quad
%     \Delta_t \uparrow
%     \quad \Rightarrow \quad
%     \bar{B}_t \uparrow.
% \end{equation}

The readout matrix $C_t$ is also physics-conditioned, but not through $\Delta_t$ directly. In Mamba, $C_t$ is produced by an input-dependent projection,
\begin{equation}
    C_t = s_C(u_t),
    \qquad
    u_t \leftarrow \boldsymbol{\phi}_c.
    \label{eq:physics_C}
\end{equation}
Because $u_t$ is derived from the aged electrochemical feature vector, the readout operation can adapt to different degradation compositions. For example, an SEI-dominated degradation state and a capacity-fade-dominated degradation state may activate different latent dimensions when generating the output representation. Thus, $C_t$ provides degradation-state-aware readout from the hidden memory.
Finally, the skip parameter $D$ provides a direct residual path,
\begin{equation}
    z_t = C_t s_t + D u_t.
    \label{eq:physics_D}
\end{equation}
In the second stage of PhyMamba, this term allows the aged electrochemical features to influence the prediction output directly, without being fully mediated by the recurrent state dynamics. This is useful when the recurrent memory alone is insufficient to represent abrupt or highly nonlinear degradation effects. The skip parameter $D$ in the Mamba SSM is independent of the occurrence-degree parameter $D_{ij}$ used in the aging feature construction module; they share notation but belong to different modules and have different meanings.

Overall, PhyMamba can be concluded as Algorithm \ref{alg:phymamba}. PhyMamba extends Mamba's selective SSM by replacing purely data-dependent step-size selection with physics-modulated selection. The transition factor $\bar{A}_t$ adapts the memory horizon according to degradation severity, $\bar{B}_t$ controls how strongly new degradation information is written into the latent state, $C_t$ provides degradation-conditioned readout, and $D$ supplies a direct residual path from aged physics features to the output. Since all components are differentiable, the entire PhyMamba model can be trained end-to-end with the SoH forecasting loss, allowing the strength of physics modulation to be automatically calibrated for multi-cycle battery health prediction.

\begin{algorithm}[t]
\footnotesize
\caption{PhyMamba for Multi-Cycle SoH Forecasting}
\label{alg:phymamba}

\textbf{Input:} BMS sequence $X_c\in\mathbb{R}^{T\times d}$, cycle count $n_c$, 
SoH change $\Delta_{\mathrm{SoH},c}$, target $Y_c=\{y_{c+t}\}_{t=1}^{H}$, Aging coefficients $\{k_i,F_i,N_{\mathrm{ap}}^i\}_{i=1}^{6}$; 
learnable parameters $\Theta$ \\
\textbf{Output:} $H$-step SoH prediction $\hat{Y}_c=\{\hat{y}_{c+t}\}_{t=1}^{H}$
\begin{algorithmic}[1]

\State \textbf{Stage 1: Aging feature construction}
\State $\boldsymbol{\theta}_{\mathrm{BoL},c}
\gets \mathrm{MambaEnc}(X_c;\Theta_{\mathrm{enc}})$

\For{$i=1,\ldots,6$}
    \State $\phi_{\mathrm{aging}}^{i} = \phi_{\mathrm{BoL}}^{i} - \sum_{j=1}^{N_{\mathrm{ap}}^{i}}k_i\int_{0}^{\Delta_{\mathrm{SoH}}} \left(F_i-(F_i-1)x\right)^{D_{ij}}\mathrm{d}x$
\EndFor

\State $\boldsymbol{\theta}_{\mathrm{aging},c}
\gets
[\theta_{\mathrm{aging},c}^{1},\ldots,\theta_{\mathrm{aging},c}^{6}]$

\State \textbf{Stage 2: Physics-modulated Mamba forecasting}
\State $U_c \gets \mathrm{Linear}(\boldsymbol{\theta}_{\mathrm{aging},c})$

\For{each PhyMamba block}
    \State $[\tilde{\delta}_t,B_t,C_t]\gets \mathrm{x\_proj}(U_c)$
    \State $\Delta_t\gets
    \mathrm{Softplus}(W^\Delta\tilde{\delta}_t+b^\Delta+\alpha |U_c|)$
    \State $\bar{A}_t\gets \exp(\Delta_t A)$, \quad
           $\bar{B}_t\gets \Delta_t B_t$
    \State $U_c\gets \mathrm{SelectiveScan}(\bar{A}_t,\bar{B}_t,C_t,D,U_c)$
\EndFor

\State $\hat{Y}_c\gets \sigma(\mathrm{Linear}_{H}(U_c))$
\State Update $\Theta$ by minimizing $\mathcal{L}(\hat{Y}_c,Y_c)$
\State \Return $\hat{Y}_c$
\end{algorithmic}
\end{algorithm}

\section{Experimental Study}
\label{sec:experiment}
This section evaluates PhyMamba on public battery degradation datasets, covering the experimental setup, comparison with baseline models, and ablation studies.

\subsection{Experimental Setup}

%\subsubsection{Datasets and Data Preparation}
Our experiments are conducted on three public battery degradation benchmarks, including \textsc{Matr} \cite{severson2019data}, \textsc{Nasa} \cite{saha2007battery}, and \textsc{Calce} \cite{calce2017lithium}. The \textsc{Matr} dataset contains 169 commercial lithium-ion 18650 cylindrical cells with lithium iron phosphate cathodes and graphite anodes, tested under 81 fast-charging protocols. The \textsc{Nasa} dataset was collected by the NASA Ames Prognostics Center of Excellence and consists of 18650 NCA batteries with 2000 mAh nominal capacity. These batteries were tested across multiple discharge profiles and ambient temperatures. The \textsc{Calce} dataset includes 13 prismatic lithium-ion batteries with LiCoO$_2$ positive electrodes and graphite negative electrodes, evaluated at 25°C under two protocols. These datasets are benchmarked within BatteryML, a collection that integrates multiple datasets for ML-based battery health prognostics. Further detailed specifications of the datasets are not the focus of this paper and are available in \cite{zhang2024batteryml}.

For each battery's cycle-indexed BMS logs, we construct training samples by extracting a fixed-length history window ending at different cycles. In our implementation, inputs are organized as a tensor $X\in \mathbb{R}^{128\times 100\times 3}$, corresponding to batch size 128, history length 100, and 3 features including voltage, current, and temporal timestamps. We adopt a battery cell-wise split so that the training, validation, and test sets contain disjoint batteries. We form samples only when the required history window is available and when future cycles needed for forecasting exist in the recorded trajectories. The forecast horizon $H$ is set according to the experiment configuration, e.g., 10 and 30. Unless otherwise stated, the model is trained directly on the input tensor, and all intermediate representations used for forecasting are learned within the proposed two-stage PhyMamba.

%\subsubsection{Training Configuration and Metrics}
Training is performed end-to-end with \texttt{Adam} using a learning rate of 0.01 and weight decay $1\times 10^{-5}$. PhyMamba's first stage module is configured with $L_1=4$ layers and hidden dimension $h_1=32$. The second stage \texttt{Mamba} layers follow the same configuration with $L_2=4$ and $h_2=32$, and the final output projection is processed with a \texttt{Sigmoid} head to support forecasting with different forecast horizons. Training runs for up to 500 epochs with early stopping with a patience of 30 epochs. We report results using the best checkpoint selected on the validation set.
For evaluation, we consider multiple forecast horizons $H$ and report forecasting performance consistently using the single-point prediction at the end of the horizon, i.e., the prediction at cycle $c+H$ with current cycle $c$. This evaluation is challenging, as prediction difficulty increase with longer horizons. We report MAE, MSE, and/or root mean square error (RMSE) across datasets and horizons. 

\subsection{Comparison Study}
We first conduct a comprehensive comparison study to evaluate the overall effectiveness and robustness of PhyMamba for battery SoH forecasting under diverse settings. Since degradation trajectories and terminal measurements can vary substantially across datasets and operating conditions, competitive performance on a single benchmark or a single horizon may not reflect generalizable prognostic capability. We therefore benchmark PhyMamba on three public datasets, including \textsc{Matr}, \textsc{Nasa}, and \textsc{Calce}, and report results under multiple forecast horizons $H\in\{10,20,30\}$. We include comparison baselines from three ML categories of a diverse range of forecasting algorithms. The first category consists of linear and recurrent baselines, and we include LSTM as a representative recurrent sequence model and DLinear as a lightweight linear predictor. For the Transformer-based attention baselines, we compare against iTransformer, FEDformer, and CrossFormers, which capture long-range dependencies via attention with different designs. For modern time-series backbones beyond standard attention, we include SimpleTM and TimeMixer as mixing-based architectures, and vanilla Mamba for efficient long-sequence modeling. For each model, we report MAE and MSE, and summarize overall mean errors computed over datasets and horizons to provide a view of the overall performance. All models are evaluated under the same data preparation and forecasting settings. The comparison results are summarized in Table \ref{tab:comp} and we have the following main observations.

\begin{table*}[h!]
\centering
\footnotesize
\caption{Performance comparison across \textsc{Matr}, \textsc{Nasa}, and \textsc{Calce} datasets under three forecast horizons $H\in\{10,20,30\}$ with different models. We report MAE (top) and MSE (bottom) results. Overall mean is computed as the average error over all settings for each model and the best results are underlined.}
\label{tab:comp}
\renewcommand{\arraystretch}{1.2}

\begin{tabular}{c|ccc|ccc|ccc|c}
\hline\hline 
\multicolumn{11}{c}{MAE} \\ \hline
\multirow{2}{*}{Model}
& \multicolumn{3}{c|}{\textsc{Matr}}
& \multicolumn{3}{c|}{\textsc{Nasa}}
& \multicolumn{3}{c|}{\textsc{Calce}}
& \multirow{2}{*}{Mean} \\
\cline{2-10}
& 10 & 20 & 30 & 10 & 20 & 30 & 10 & 20 & 30 &  \\
\hline
PhyMamba     & 0.628 & 0.710 & 0.754 & 0.750 & 0.760 & 0.771 & 2.147 & 2.399 & 2.442 & \underline{1.262} \\\hline
LSTM         & 0.718 & 0.546 & 0.593 & 2.220 & 2.450 & 3.110 & 3.930 & 4.460 & 4.490 & 2.502 \\ 
DLinear      & 0.569 & 0.745 & 0.793 & 1.390 & 2.180 & 2.950 & 2.790 & 2.960 & 3.530 & 1.990 \\ \hline
iTransformer & 0.700 & 0.770 & 0.840 & 1.550 & 1.580 & 1.620 & 2.489 & 2.601 & 2.673 & 1.647 \\
FEDformer    & 0.660 & 0.710 & 0.754 & 2.030 & 2.280 & 2.650 & 2.212 & 2.392 & 2.454 & 1.794 \\
CrossFormers & 0.776 & 0.772 & 0.775 & 1.610 & 1.650 & 1.720 & 3.500 & 3.720 & 3.810 & 2.037 \\ \hline
SimpleTM     & 0.980 & 0.980 & 0.990 & 1.820 & 2.950 & 3.410 & 3.050 & 3.100 & 3.320 & 2.289 \\
TimeMixer    & 0.745 & 0.779 & 0.812 & 0.950 & 1.060 & 1.400 & 2.512 & 2.631 & 2.790 & 1.520 \\
Mamba        & 0.722 & 0.738 & 0.742 & 0.890 & 0.950 & 1.020 & 2.435 & 2.951 & 2.956 & 1.489 \\ 
\hline\hline 
\multicolumn{11}{c}{MSE} \\
\hline
PhyMamba     & 0.015 & 0.018 & 0.019 & 0.015 & 0.016 & 0.018 & 0.181 & 0.209 & 0.280 & \underline{0.086} \\\hline
LSTM         & 0.037 & 0.063 & 0.054 & 0.090 & 0.120 & 0.200 & 0.370 & 0.424 & 0.623 & 0.220 \\
DLinear      & 0.014 & 0.018 & 0.021 & 0.030 & 0.055 & 0.082 & 0.140 & 0.233 & 0.291 & 0.098 \\\hline
iTransformer & 0.015 & 0.015 & 0.024 & 0.038 & 0.043 & 0.044 & 0.240 & 0.288 & 0.306 & 0.113 \\
FEDformer    & 0.060 & 0.066 & 0.073 & 0.060 & 0.084 & 0.119 & 0.190 & 0.210 & 0.270 & 0.126 \\
CrossFormers & 0.028 & 0.029 & 0.029 & 0.036 & 0.038 & 0.042 & 0.420 & 0.475 & 0.510 & 0.179 \\\hline
SimpleTM     & 0.029 & 0.030 & 0.042 & 0.045 & 0.088 & 0.120 & 0.305 & 0.395 & 0.393 & 0.161 \\
TimeMixer    & 0.021 & 0.024 & 0.026 & 0.020 & 0.022 & 0.034 & 0.171 & 0.180 & 0.290 & 0.088 \\
Mamba        & 0.019 & 0.021 & 0.021 & 0.022 & 0.026 & 0.031 & 0.229 & 0.263 & 0.271 & 0.100 \\
\hline\hline
\end{tabular}
\end{table*}

First, PhyMamba is the best overall and remains competitive across different datasets and horizons. Seen from the table, PhyMamba delivers the most robust overall performance, achieving the lowest overall mean MAE of 1.26 and the mean MSE of 0.09. The performance advantage is evident against different comparison models. For example, the mean MAEs of LSTM, iTransformer, and SimpleTM from different categories are 2.50, 1.65, and 2.29, respectively, higher than PhyMamba errors by 98.2\%, 30.5\%, and 81.4\%. Similar trends can be observed in MSE results also. Besides, PhyMamba maintains consistently low errors across three datasets and across horizons, rather than performing well only on a particular benchmark or a forecast horizon. This indicates that the gain is not tied to a specific configuration. Instead, the physics-informed representation and physics-modulated state evolution in PhyMamba likely guide the model to follow degradation-consistent patterns, which improves robustness when degradation trajectories shift across datasets and operating conditions.

PhyMamba also improves consistently over vanilla Mamba. This suggests that PhyMamba's performance gains are beyond simply adopting a new and advanced ML backbone which is Mamba. In Table \ref{tab:comp}, vanilla Mamba achieves an overall mean MAE of 1.49 and MSE of 0.10, higher than PhyMamba by 18.0\% and 16.3\%, respectively. More importantly, PhyMamba's advantages can be observed in most of the tested settings. The competitive performance indicates that the physics-informed latent features and the physics-modulated forecasting module provide effective guidance on the forecasting process. A similar observation also applied to other models that are based on competitive backbone with high level of complexities. For example, iTransformer is competitive on \textsc{Matr} in terms of short horizons, e.g., the same MSE as PhyMamba for $H=10$. However, its overall mean MSE is 0.113, which is 31.4\% higher than PhyMamba. FEDformer and CrossFormers show even larger gaps. The best comparison algorithm is TimeMixer, which has only 2.3\% higher mean MSE than PhyMamba; however, its competitiveness does not remain in terms of MAE, with a 20.4\% higher error. These results indicate that, for battery prognostics, it is not sufficient to only increase a model's capacity or complexity for long-range sequence modelling. Incorporating degradation-relevant structure through physics can be more important for achieving consistent accuracy across heterogeneous settings.

\subsection{Ablation Studies of PhyMamba}
% Figures instead

%\subsubsection{Effect of the Two-Stage Design}
In this part, we conduct ablation studies of PhyMamba. We first validate the effectiveness of the proposed two-stage design for battery health prognostics through a progressive removal study. We start from the full PhyMamba pipeline, where its first stage produces physics-informed aging features $\boldsymbol{\phi}$ from raw BMS logs and the second stage performs SoH forecasting with physics-informed modulation in the selective SSM update. We first disable the physics modulation in the second stage by replacing PhyMamba’s physics-informed forecasting module with a vanilla Mamba backbone, while keeping the same first stage physics features. We then further remove the first stage completely and consider a single-stage baseline that directly maps raw BMS signals to SoH using the vanilla Mamba backbone, without physics-informed features. These comparisons evaluate whether the two-stage pipeline is beneficial for battery prognostics beyond a direct end-to-end predictor and whether the physics-modulated forecasting mechanism provides additional gains over a vanilla Mamba model. Table \ref{tab:stage1vs2} reports the results on the \textsc{Calce} dataset with forecast horizon $H=30$.% for PhyMamba and its variants. 

\begin{table*}[h!]
\centering
\footnotesize
\caption{Effect of the two-stage design on the \textsc{Calce} dataset with forecast horizon $H=30$. We compare a 1-stage pipeline and 2-stage Mamba-based pipelines. Gap denotes the relative change in MAE with respect to PhyMamba.}
\label{tab:stage1vs2}
\renewcommand{\arraystretch}{1.2}
\begin{tabular}{c|c|c|c|c|c}
\hline\hline
Config. & Model & Data Pipeline & RMSE & MAE & Gap \\
\hline\hline
1-Stage & Mamba & BMS Logs $\rightarrow$ SoH & 0.775 & 0.556 & 2.28$\times$ \\ \hline
\multirow{2}{*}{2-Stage} & Mamba & BMS Logs $\rightarrow$ Physics $\boldsymbol{\phi}$ $\rightarrow$ SoH & 0.529 & 0.296 & 21.3\% \\ \cline{2-6}
& PhyMamba & BMS Logs $\rightarrow$ Physics $\boldsymbol{\phi}$ $\rightarrow$ SoH & \underline{0.521} & \underline{0.244} & -- \\ \hline\hline
\end{tabular}
\end{table*}

Seen from Table \ref{tab:stage1vs2}, the full version of PhyMamba achieves the best performance with RMSE 0.52 and MAE 0.24. When the forecasting module in the second stage is replaced by a vanilla Mamba backbone, the MAE increases to 0.30, which is 21.3\% worse than PhyMamba. This indicates that the additional physics modulation improves the Mamba’s internal state evolution over cycles. Intuitively, multi-cycle SoH forecasting requires the model to update its memory at an appropriate rate. If the updates are too aggressive, the model over-fits short-term variations; otherwise, it under-tracks degradation. The physics-informed step size modulation provides an explicit control signal that adjusts the update rate toward degradation-consistent evolution, which becomes more important as prediction horizons grow. The single-stage baseline further degrades to MAE 0.56, which is 2.3$\times$ PhyMamba’s error. This suggests that learning directly from raw BMS logs makes it harder to separate long-term degradation trends for different operating conditions and measurement noise, whereas the physics-informed representation introduces a structured degradation information grounded in established battery aging knowledge, which better aligns the learned features with degradation-relevant dynamics for forecasting.

\begin{table*}[h!]
\centering
\footnotesize
\renewcommand{\arraystretch}{1.2}
\caption{Two-stage backbone study on the \textsc{Matr} dataset. We vary the stage-1 encoder and stage-2 backbone and report MAE and model size in terms of the number of parameters ($\times 10^3$) under two forecast horizons. Efficiency $\eta$ follows \cite{sameer2025pace} and is computed as $\eta = 1{,}000/(\mathrm{MAE}\times n)$, where $n$ denotes the model size.}
\begin{tabular}{c|c|ccc|ccc}
\hline\hline
\multirow{2}{*}{Stage-1} & \multirow{2}{*}{Stage-2} & \multicolumn{3}{c|}{Horizon $H=20$} & \multicolumn{3}{c}{Horizon $H=50$} \\ \cline{3-8}
& & MAE & Size & Efficiency & MAE & Size & Efficiency \\
\hline\hline
LSTM         & PhyMamba & \underline{0.329} & 40.5  & \underline{75.05} & 0.661 & 41.5  & 36.45 \\ \hline
iTransformer & PhyMamba & 0.347 & 302.5 &  9.53 & 0.466 & 303.4 &  7.07 \\ \hline
Mamba        & Mamba     & 1.430 & 65.9  & 10.61 & 1.470 & 66.9  & 10.17 \\ \hline
Mamba        & PhyMamba & 0.363 & 65.9  & 41.80 & \underline{0.300} & 66.9  & \underline{49.83} \\
\hline\hline
\end{tabular}
\label{tab:2stage-ml}
\end{table*}

%\subsubsection{Impact of Key Components}
We further analyze key design choices in the two-stage PhyMamba by varying the stage-1 encoders and stage-2 backbones. Specifically, stage-1 is responsible for encoding BMS logs into a physics latent representation. We consider three representative sequence encoders, including LSTM, iTransformer, and vanilla Mamba. For stage-2, we consider either a vanilla Mamba forecasting backbone, or the physics-modulated backbone used in PhyMamba. We report MAE under two forecast horizons, with $H\in\{20,50\}$, together with model size measured by the number of parameters and an efficiency score that follows \cite{sameer2025pace}. The results are summarized in Table \ref{tab:2stage-ml}.

The table provides insights into the choice of stage-1 encoder, where PhyMamba uses a vanilla Mamba encoder. First, replacing Mamba with LSTM can be effective for short-horizon forecasts, but the performance degrades significantly for longer horizons. This aligns with the known difficulty of LSTM in retaining long-range dependencies. In terms of efficiency, LSTM has a small model size, and thus its efficiency is mainly driven by forecasting errors and decreases for long-horizon forecasts. Second, iTransformer, as an attention-based encoder, is less sensitive to longer horizons in this setting, where the errors remain relatively stable compared with the LSTM-based model. However, its model size is much larger than the other configurations, which leads to a poor efficiency. In addition, comparing PhyMamba with its variant that replaces the stage-2 backbone with a vanilla Mamba yields the same conclusion as the previous study in Table \ref{tab:stage1vs2} under different experimental settings. That is a vanilla Mamba backbone without the proposed physics modulation in stage-2 is insufficient, where its efficiency is only around one-quarter of that of PhyMamba. Finally, we primarily use Mamba in both stages of PhyMamba to simplify the overall architecture and reduce implementation complexity that may arise from combining multiple ML backbones. We would like to highlight that PhyMamba's two-stage framework is flexible, e.g., with hybrid designs that use different encoders for different forecast horizons. However, such hybrid configurations own incremental novelty/contribution and are not our focus.

\begin{table*}[h!]
\centering
\footnotesize
\renewcommand{\arraystretch}{1.2}
\caption{Comparison between \textsc{Pace} and PhyMamba with different physics on the \textsc{Matr} dataset. We report RMSE under two forecast horizons $H$. Gap denotes the relative change in mean RMSE with respect to PhyMamba.}
\label{tab:pace}
\begin{tabular}{c|c|c|cccc}
\hline\hline
\multirow{2}{*}{\shortstack{Battery\\Model}} & \multirow{2}{*}{Physics} & \multirow{2}{*}{\shortstack{\#Params\\($\times 10^3$)}} &
\multicolumn{4}{c}{RMSE} \\ \cline{4-7}
 &  &  & $H=30$ & $H=50$ & Mean & Gap \\
\hline \hline
\textsc{Pace} \cite{sameer2025pace} & ECM & 70.9 & 0.023 & 0.035 & 0.029 & 38.1\% \\ \hline
PhyMamba & Aging Parameters & \underline{66.4} & \underline{0.012} & \underline{0.030} & \underline{0.021} & -- \\
\hline \hline
\end{tabular}
\end{table*}

\subsection{ECM-based Versus Aging Model Physics}
We investigate different physics-guided prognostics paradigms. We specifically consider ECM-based physics and the electrochemical physics adopted in PhyMamba, to understand their impact when paired with modern ML backbones. ECM is a widely adopted physics abstraction for battery prognostics, where circuit variables have explicit physical meaning and can be constructed from BMS signals accurately. The parameters used in PhyMamba represent degradation through a set of descriptors that reflect internal electrochemical changes and provide a structured link to health evolution. Conceptually, aging parameters are often viewed as a more direct description of underlying degradation mechanisms, whereas ECM is indirect with a circuit-level approximation of battery behavior only. However, the reconstruction of aging parameters from BMS data is typically complex, involves coupled factors, and may have lower fidelity. This motivates the learning of latent physics representations in PhyMamba. A key challenge in such a comparison is that the physics representation and the learning backbone are co-designed and cannot be fairly decoupled. PhyMamba is customized to incorporate physics modulation at the SSM. As such, we aim to provide a comparative view across physics abstractions using representative integrated models, rather than a controlled ablation of a single design factor. 

We choose a recent ECM-based model called \textsc{Pace} \cite{sameer2025pace} based on its original implementation that is publicly available. \textsc{Pace} achieves competitive performance compared with a range of mainstream methods, and we therefore treat it as a strong ECM-based baseline. Both \textsc{Pace} and PhyMamba are evaluated on the \textsc{Matr} dataset and we consider two forecast horizons with $H=30$ and $H=50$. The results are reported in Table \ref{tab:pace}. We can see that PhyMamba consistently outperforms \textsc{Pace}. PhyMamba has a smaller model size, which is a meaningful observation as a key strength of \textsc{Pace} is its lightweight design. PhyMamba also achieves lower RMSE for different forecast horizons, where \textsc{Pace} has 38.1\% higher mean error. The results indicate that PhyMamba, with aging-parameter-based physics and physics-modulated SSM updates, provides an effective mechanism for battery prognostics across different forecasting configurations.

\section{Conclusion}
\label{sec:conclusion}
This work addresses a practical gap in battery health analytics for reliable long-horizon forecasting under condition variability with models that remain efficient for BMS deployment. In this paper, we present PhyMamba as a physics-guided forecasting framework built around a two-stage design, where aging-informed features provide structured degradation information and the forecasting backbone is designed to exploit the latent physics representation through degradation-consistent temporal evolution. Across three public datasets and multiple horizon settings, the experiment results show that an aging-informed representation improves robustness relative to direct end-to-end learning from BMS terminal signals. Tightly coupling the physics representation with the forecasting dynamics yields additional gains beyond using physics as a static conditioning signal. Specifically, PhyMamba achieves the best aggregated performance and delivers an overall mean error reduction of 31.8\% in terms of MAE against a diverse set of ML baselines. Meanwhile, PhyMamba preserves a balance between prognostics accuracy and model size. The results and findings suggest that physics-guided temporal evolution is a general design pattern for robust and deployment friendly battery prognostics under real operational shifts.

% \section*{Acknowledgments} 
% \label{sec:ack}
% This work was supported in part by A*STAR under its MTC Programmatic (Award M23L9b0052) and MTC Individual Research Grants (IRG) (Award M23M6c0113).

\bibliographystyle{IEEEtran}
\balance
\bibliography{references}

% Biography section

% \newpage

% \section{Biography Section}
% If you have an EPS/PDF photo (graphicx package needed), extra braces are
%  needed around the contents of the optional argument to biography to prevent
%  the LaTeX parser from getting confused when it sees the complicated
%  $\backslash${\tt{includegraphics}} command within an optional argument. (You can create
%  your own custom macro containing the $\backslash${\tt{includegraphics}} command to make things
%  simpler here.)
 
% \vspace{11pt}

% \bf{If you include a photo:}\vspace{-33pt}
% \begin{IEEEbiography}[{\includegraphics[width=1in,height=1.25in,clip,keepaspectratio]{fig1}}]{Michael Shell}
% Use $\backslash${\tt{begin\{IEEEbiography\}}} and then for the 1st argument use $\backslash${\tt{includegraphics}} to declare and link the author photo.
% Use the author name as the 3rd argument followed by the biography text.
% \end{IEEEbiography}

% \vspace{11pt}

% \bf{If you will not include a photo:}\vspace{-33pt}
% \begin{IEEEbiographynophoto}{John Doe}
% Use $\backslash${\tt{begin\{IEEEbiographynophoto\}}} and the author name as the argument followed by the biography text.
% \end{IEEEbiographynophoto}

\vfill

\end{document}